\documentclass[sigconf]{acmart}
\RequirePDFTeX

\usepackage{comment}
\usepackage{listings}
\usepackage{tabularx}
\usepackage{booktabs}
\usepackage{siunitx}
\usepackage{enumitem}
\usepackage{makecell}

\AtBeginDocument{%
  \providecommand\BibTeX{{%
    \normalfont B\kern-0.5em{\scshape i\kern-0.25em b}\kern-0.8em\TeX}}}

\copyrightyear{2024}
\acmYear{2024}
\setcopyright{cc}
\setcctype[4.0]{by}
\acmConference[GECCO '24 Companion]{Genetic and Evolutionary Computation Conference}{July 14--18, 2024}{Melbourne, VIC, Australia}
\acmBooktitle{Genetic and Evolutionary Computation Conference (GECCO '24 Companion), July 14--18, 2024, Melbourne, VIC, Australia}
\acmDOI{10.1145/3638530.3664144}
\acmISBN{979-8-4007-0495-6/24/07}

\begin{document}

\title{Investigating Hyperparameter Optimization and Transferability for ES-HyperNEAT: A TPE Approach}
\titlenote{Corrected version (7 September 2026) of the paper published in the
GECCO~'24 Companion proceedings, \url{https://doi.org/10.1145/3638530.3664144}.
It fixes typographical errors, the header grouping of
Table~\ref{tab:best_hyperparameters_compact}, and the affiliation names; all results
and claims are unchanged.}

\author{Romain Claret}
\email{romain.claret@unine.ch}
\orcid{0000-0002-5612-8471}
\affiliation{%
  \institution{University of Neuchâtel}
  \streetaddress{Information Management Institute, A.L.Breguet 2}
  \city{Neuchâtel}
  \country{Switzerland}
  \postcode{2000}
}

\author{Michael O'Neill}
\email{m.oneill@ucd.ie}
\orcid{0000-0001-8734-417X}
\affiliation{%
  \institution{University College Dublin}
  \streetaddress{UCD Natural Computing Research \& Applications Group}
  \city{Dublin}
  \country{Ireland}
  \postcode{A94 XF34}
}

\author{Paul Cotofrei}
\email{paul.cotofrei@unine.ch}
\orcid{0000-0002-4103-5467}
\affiliation{%
  \institution{University of Neuchâtel}
  \streetaddress{Information Management Institute}
  \city{Neuchâtel}
  \country{Switzerland}
  \postcode{2000}
}

\author{Kilian Stoffel}
\email{kilian.stoffel@unine.ch}
\orcid{0000-0002-9486-7769}
\affiliation{%
  \institution{University of Neuchâtel}
  \streetaddress{Information Management Institute}
  \city{Neuchâtel}
  \country{Switzerland}
  \postcode{2000}
}

\renewcommand{\shortauthors}{Claret and O'Neill, et al.}

\begin{abstract}
Neuroevolution of Augmenting Topologies (NEAT) and its advanced version, Evolvable-Substrate HyperNEAT (ES-HyperNEAT), have shown great potential in developing neural networks. However, their effectiveness heavily depends on the selection of hyperparameters. This study investigates the optimization of ES-HyperNEAT hyperparameters using the Tree-structured Parzen Estimator (TPE) on the MNIST classification task, exploring a search space of over 3 billion potential combinations. TPE effectively navigates this vast space, significantly outperforming random search in terms of mean, median, and best accuracy. During the validation process, the best hyperparameter configuration found by TPE achieves an accuracy of 29.00\% on MNIST, surpassing previous studies while using a smaller population size and fewer generations. The transferability of the optimized hyperparameters is explored in logic operations and Fashion-MNIST tasks, revealing successful transfer to the more complex Fashion-MNIST problem but limited to simpler logic operations. This study emphasizes a method to unlock the full potential of neuroevolutionary algorithms and provides insights into the hyperparameters' transferability across tasks of varying complexity.
\end{abstract}



\received{8 April 2024}
\received[revised]{10 May 2024}
\received[accepted]{3 May 2024}

\maketitle

\hyphenation{emer-ged}
\hyphenation{Topolo-gies}

\section{Introduction}
Neuroevolution, a subfield of evolutionary computation, emerged as a powerful approach for optimizing neural networks, with algorithms like Neuroevolution of Augmenting Topologies (NEAT)~\cite{stanley2002evolving} and its extensions demonstrating promising results across various domains. The performance of these algorithms heavily relies on the selection of hyperparameters, which significantly impact the search process and the resulting network architectures~\cite{risi2010indirectly}. This paper investigates hyperparameter optimization for the Evolvable-Substrate HyperNEAT (ES-HyperNEAT) algorithm~\cite{risi2012enhanced}, an indirect encoding extension of NEAT, using the Tree-structured Parzen Estimator (TPE)~\cite{bergstra2011algorithms} on the MNIST classification task~\cite{lecun1998mnist}.

TPE efficiently explores the search space and identifies promising configurations by modeling the probability distribution of well-performing hyperparameters. This study assesses TPE's potential for optimizing ES-HyperNEAT hyperparameters and investigates the transferability of optimized configurations to tasks of varying complexity, such as logic operations and Fashion-MNIST classification~\cite{xiao2017fashion}.

The successful application of ES-HyperNEAT to various tasks, such as logic operations~\cite{risi2012enhanced}, has been well-documented. However, the MNIST classification task remains a challenge for the algorithm in its pure form, with a study by Verbancsics and Harguess~\cite{verbancsics2015image} achieving an accuracy of 23.90\% using HyperNEAT with a predefined LeNet-5 topology~\cite{lecun1998gradient}.

This research has significant potential to impact the field of neuroevolution and its applications. By laying the foundations for more efficient and adaptable neuroevolutionary algorithms, the insights gained can guide the development of robust and generalizable neuroevolutionary systems, enabling their application to a broader range of complex real-world problems. The remainder of this paper is organized as follows: Section 2 provides background information on ES-HyperNEAT and TPE, Section 3 presents an overview of related work, Section 4 describes the experimental setup, Section 5 presents the experimental results and discusses the effectiveness of TPE optimization and the transferability of the best hyperparameter configuration, Section 6 discusses the findings' implications, limitations, and potential future research directions, and finally, Section 7 concludes the paper, summarizing the main contributions and significance of the research.

\section{Background}
\subsection{Neuroevolution and ES-HyperNEAT}
Neuroevolution is a subfield of evolutionary computation that applies evolutionary algorithms to optimize neural networks. The Neuroevolution of Augmenting Topologies (NEAT) algorithm~\cite{stanley2002evolving} evolves both the weights and the structure of neural networks, starting with a minimal network and incrementally adding nodes and connections through mutation and crossover operations inspired by natural evolution principles.

Hypercube-based NEAT (HyperNEAT)~\cite{stanley2009hypercube} extends NEAT by evolving Compositional Pattern Producing Networks (CPPNs)~\cite{stanley2007compositional}, which generate the connectivity patterns and weights of a fixed-topology substrate network. HyperNEAT exploits the geometric regularities of the problem domain to evolve large-scale neural networks with complex connectivity patterns. Evolvable-Substrate HyperNEAT (ES-HyperNEAT)~\cite{risi2012enhanced} further extends HyperNEAT by allowing the substrate network's topology to evolve along with the CPPN, enabling the discovery of more efficient and task-specific network topologies.

\subsection{Bayesian Optimization and TPE}
Bayesian optimization~\cite{shahriari2015taking} is a sequential model-based optimization approach well-suited for expensive black-box functions, such as those encountered in hyperparameter optimization. It constructs a probabilistic model of the objective function to guide the search towards promising regions of the search space.

The Tree-structured Parzen Estimator (TPE)~\cite{bergstra2011algorithms} is a Bayesian optimization algorithm that models the probability distribution of hyperparameters that lead to good performance. TPE maintains two density estimators: $l(\mathbf{x})$ for the distribution of hyperparameters that yield low objective values and $g(\mathbf{x})$ for the distribution of hyperparameters that yield high objective values.

At each iteration, TPE selects the next set of hyperparameters to evaluate by maximizing the expected improvement (EI) acquisition function:

$$\text{EI}(\mathbf{x}) = \int_{-\infty}^{y^*} (y^* - y) \frac{l(\mathbf{x})}{g(\mathbf{x})} p(y | \mathbf{x}) \, dy$$

\noindent where $y^*$ is the current best observed objective value, and $p(y | \mathbf{x})$ is the probability of observing an objective value $y$ given the hyperparameters $\mathbf{x}$. TPE efficiently explores the search space by iteratively updating the density estimators and selecting hyperparameters that maximize the expected improvement.

\section{Related Work}
Neuroevolution has emerged as a powerful approach to optimizing neural networks, with the NEAT algorithm~\cite{stanley2002evolving} and its extensions demonstrating promising results across various domains. A systematic literature review by Papavasileiou et al.~\cite{papavasileiou2021systematic} categorizes these developments based on their key contributions and application areas, highlighting the diverse range of enhancements made to NEAT, such as the incorporation of indirect encoding schemes, the evolution of modular and hierarchical structures, and the application of neuroevolution to domains like deep learning, reinforcement learning, and robotics.

The importance of hyperparameters in NEAT and its extensions has been well-established. Stanley and Miikkulainen~\cite{stanley2002evolving} provided guidelines for setting these hyperparameters based on the problem domain. Risi and Stanley further investigated the impact of hyperparameters on HyperNEAT's performance~\cite{risi2010indirectly}. In the original HyperNEAT paper, Stanley et al.~\cite{stanley2009hypercube} solved the XOR logic operation using hand-tuned hyperparameters. Later, Risi and Stanley~\cite{risi2012enhanced} showcased ES-HyperNEAT's capabilities across various tasks using hand-tuned hyperparameters. While these hand-tuned hyperparameters yielded impressive results, their work highlighted the potential for further improvement through systematic hyperparameter optimization.

In the field of hyperparameter optimization, Bayesian optimization approaches have gained popularity due to their ability to explore high-dimensional search spaces efficiently. TPE~\cite{bergstra2011algorithms} has shown promising results in various machine learning tasks, demonstrating its effectiveness in handling high-dimensional hyperparameter spaces~\cite{bergstra2011algorithms, snoek2012practical}.

While these studies provide valuable insights into hyperparameter optimization and TPE's effectiveness, more research is needed, focusing specifically on applying TPE to ES-HyperNEAT and exploring the transferability of optimized hyperparameters from a specific source task to another target task with lower or similar complexity. Our work aims to fill this gap by investigating the use of TPE for optimizing ES-HyperNEAT's hyperparameters, exploring the transferability of the optimized hyperparameters from the MNIST classification task, selected as the source task, to logic operations and Fashion-MNIST classification tasks, chosen as the target tasks, and highlighting the potential for further advancements in the field.

\section{Experimental Setup}

Our study comprises two main experiments: hyperparameter investigation and transferability of hyperparameters. The hyperparameter investigation evaluates TPE's~\cite{bergstra2011algorithms} effectiveness in finding performant ES-HyperNEAT~\cite{risi2012enhanced} hyperparameter configurations on the MNIST classification task~\cite{lecun1998mnist}, comparing its performance to random search~\cite{bergstra2012random}. The transferability experiment examines the generalizability of the best MNIST hyperparameter configuration to logic operations and Fashion-MNIST classification tasks~\cite{xiao2017fashion}.

The experiments were implemented using Python and the Pureples framework~\cite{westh2017pureples}, which is based on NEAT-Python~\cite{McIntyre_neat-python}, and were conducted exclusively on CPUs across 13 machines with varying CPU performances and RAM sizes using the Optuna framework~\cite{akiba2019optuna}.

\subsection{Hyperparameter Investigation}

We selected sixteen hyperparameters from the ES-HyperNEAT algorithm to create a search space of over 3 billion potential configurations (tab~\ref{tab:hyperparameters}). For the hyperparameters not included in the search space, we used the configuration file from the XOR problem experiment in the NEAT-Python library and left all other parameters at the library's default values. Although TPE has been demonstrated on a 32-dimensional problem~\cite{bergstra2011algorithms}, we followed the rule of thumb in Frazier's tutorial on Bayesian Optimization~\cite{frazier2018tutorial}, which suggests keeping the number of dimensions below 20 for efficacy. Based on a preliminary study, we determined the search space, number of generations, and batch size. Both TPE and random search were employed to explore this space. The TPE search was constrained to 78 days, with each hyperparameter configuration running for 20 generations.

\begin{table}[htbp]
    \centering
    \footnotesize
    \begin{tabularx}{\columnwidth}{lX}
    \toprule
        Hyperparameter & Range of Options \\
    \midrule
        \multicolumn{2}{l}{\textbf{NEAT Parameters}} \\
        num\_hidden & 0, 1, 2, 3, 4, 5 \\
        pop\_size & 10, 20, 40, 60, 80, 100 \\
        conn\_add\_prob & 0.3, 0.5, 0.7 \\
        conn\_delete\_prob & 0.3, 0.5, 0.7 \\
        node\_add\_prob & 0.2, 0.4, 0.6, 0.8 \\
        node\_delete\_prob & 0.2, 0.4, 0.6, 0.8 \\
        initial\_connection & full\_nodirect, full\_direct, unconnected, fs\_neat\_hidden, fs\_neat\_nohidden \\
        connection\_fraction & 0.2, 0.4, 0.6, 0.8 \\
        activation\_default & sigmoid, gauss, clamped, relu, tanh, softplus \\
    \midrule
        \multicolumn{2}{l}{\textbf{Substrate Parameters}} \\
        initial\_depth & 1, 2 \\
        max\_depth & 3, 4, 5 \\
        variance\_threshold & 0.01, 0.02, 0.03, 0.04 \\
        division\_threshold & 0.3, 0.5, 0.7 \\
        max\_weight & 3.0, 5.0, 7.0 \\
        iteration\_level & 0, 1, 2, 3 \\
        activation & sigmoid, gauss, clamped, relu, tanh, softplus \\
    \bottomrule
    \end{tabularx}
    \caption{Hyperparameter search space for the Neuroevolution of Augmenting Topologies (NEAT) algorithm and substrate parameters.}
    \label{tab:hyperparameters}
\end{table}

For the MNIST experiment, the substrate was initialized at the center of the input and output vectors. At each generation, a batch of 200 random images from the MNIST dataset, evenly spread across the classes, is selected for evaluation. The fitness of each genome was evaluated by comparing the predicted class of the phenotype network with the ground truth class for each input image in the data batch. The predicted class was determined by the index of the maximum value in the phenotype's output vector, while the ground truth class was determined by the index of the maximum value in the ground truth vector. The genotype's fitness score increased by 1 for each correct prediction, and the overall fitness was calculated as the mean fitness across all evaluated images. The objective value for TPE optimization was the best fitness achieved by the best-performing individual in the final generation.

\subsection{Transferability of Hyperparameters}

The transferability experiment investigates the generalizability of the best hyperparameter configuration found during the TPE search on the MNIST task to logic operations (XOR, OR, AND, NOR, XNOR, and NAND) and Fashion-MNIST classification tasks. The motivation is to assess the effectiveness of the optimized hyperparameters in solving tasks with varying complexity and explore the potential for leveraging knowledge gained from one task to improve performance on other tasks.

For the logic operation tasks, which are solved problems, we evaluate the performance of ES-HyperNEAT using the best hyperparameter configuration from the MNIST task and compare it to a random search. The fitness for each genome is evaluated by comparing the network's output with the expected output for all possible input combinations, using 1 - the residual sum of squares (RSS)~\cite{stanley2002evolving}. Based on a preliminary study on XOR, we fixed the number of generations to 10 for each trial for each logic operation task.

For the Fashion-MNIST classification task, which presents a similar but more complex challenge compared to the MNIST task~\cite{xiao2017fashion}, we first establish a baseline performance using random search and then apply the best hyperparameter configuration from the MNIST task to assess its transferability. The substrate initialization, fitness evaluation, and selection of 200 random images per generation follow the same approach as the MNIST task, with 20 generations per trial.

We employ several performance metrics, including mean accuracy, median accuracy, best accuracy, worst accuracy, and standard deviation, to assess the performance of TPE optimization compared to random search and evaluate the transferability of the best hyperparameter configuration. We conduct two-sample t-tests, assuming unequal variances~\cite{welch1947generalization}, to determine the statistical significance of the differences between TPE and random search, as well as the transferred configuration and random search. Additionally, we calculate Cohen's d, a measure of effect size, to quantify the magnitude of the differences between the methods~\cite{cohen2013statistical}.

\section{Experimental Results}

This section presents the results of our study on hyperparameter optimization for the ES-HyperNEAT algorithm using the Tree-structured Parzen Estimator (TPE) and the transferability of the best hyperparameter configuration across different tasks. We evaluate the performance of TPE compared to random search on the MNIST classification task, investigate the transferability of the best hyperparameter configuration to logic operations and Fashion-MNIST classification tasks, and discuss the validation of our results.

\subsection{Evaluating TPE on MNIST Classification}
\label{subsec:tpe}

We compared the TPE and random search experiments to assess TPE's effectiveness for optimizing ES-HyperNEAT hyperparameters on the MNIST classification task. The random search experiment consists of at least 30 trials, a commonly used sample size for the Central Limit Theorem~\cite{devore2016probability}. However, to investigate whether TPE exhibits learning or behaves similarly to random search, we extended the number of random search trials to 292. The comparison between RandomSearch-292 and RandomSearch-30 yields a t-statistic of 0.11, a p-value of 0.910, and a Cohen's d effect size of 0.02 (tab~\ref{tab:mnist_tpe_random_two_sample}), indicating no significant difference between the performance of random search with 30 and 292 trials. Based on these results, we use RandomSearch-292 in our analysis while leaving RandomSearch-30 in the tables. This approach allows a concise presentation of the findings without compromising the validity of the conclusions drawn from the comparative analysis between random search and TPE optimization.

Table~\ref{tab:mnist_tpe_random_accuracies} presents the descriptive statistics for RandomSearch-292, TPE-Search with 2013 trials, and the best hyperparameter configuration found by TPE (TPE-Best) on the MNIST classification task, validated over 30 independent runs. The results demonstrate that TPE-Search consistently outperforms RandomSearch-292 regarding mean, median, and best accuracy values. TPE-Best achieves the highest performance, reaching an accuracy of 28.0\% during validation, surpassing the best accuracy of TPE-Search (27.5\%) while using a significantly smaller population size and fewer generations than the HyperNEAT study by Verbancsics and Harguess~\cite{verbancsics2015image} (population size of 256 and 2500 generations), which reached 23.90\%. This result highlights the potential for building more computationally efficient topologies by identifying the optimal hyperparameters.

\begin{table}[htbp]
    \centering
    \footnotesize
    \begin{tabularx}{\columnwidth}{lSSSSS}
    \toprule
        Method & {Mean (\%)} & {Median (\%)} & {SD (\%)} & {Best (\%)} & {Worst (\%)} \\
    \midrule
        RandomSearch-30 & 11.47 & 10.00 & 2.90 & 20.50 & 10.00 \\
        RandomSearch-292 & 11.41 & 10.00 & 2.76 & 20.50 & 10.00 \\
        TPE-Search & 17.41 & 19.00 & 4.26 & 27.50 & 10.00 \\
        TPE-Best & 23.40 & 23.25 & 2.07 & 28.00 & 20.00 \\
    \bottomrule
    \end{tabularx}
    \caption{Accuracy metrics for the MNIST classification task using random search with 30 and 292 trials, TPE optimization search, and the best hyperparameter configuration found by TPE.}
    \label{tab:mnist_tpe_random_accuracies}
\end{table}

To quantify the significance of the observed differences, we conducted two-sample t-tests, assuming unequal variances (tab~\ref{tab:mnist_tpe_random_two_sample}). The results reveal highly significant differences between TPE-Search and RandomSearch-292, TPE-Best and TPE-Search, and TPE-Best and RandomSearch-292, with large effect sizes. These findings emphasize the practical significance of the performance improvements achieved by TPE optimization over random search and the superiority of the best hyperparameter configuration found by TPE.

\begin{table}[htbp]
    \centering
    \footnotesize
    \begin{tabularx}{\columnwidth}{XSSS}
        \toprule
        Comparison & {t-statistic} & {p-value} & {Cohen's d} \\
        \midrule
        RandomSearch-292 vs. RandomSearch-30 & 0.11 & 0.912 & 0.02 \\
        TPE-Best vs. TPE-Search & 15.88 & {$<$0.001} & 1.42 \\
        TPE-Best vs. RandomSearch-30 & 18.86 & {$<$0.001} & 4.87 \\
        TPE-Best vs. RandomSearch-292 & 29.91 & {$<$0.001} & 4.34 \\
        TPE-Search vs. RandomSearch-30 & 11.31 & {$<$0.001} & 1.41 \\
        TPE-Search vs. RandomSearch-292 & 31.47 & {$<$0.001} & 1.47 \\
        \bottomrule
    \end{tabularx}
    \caption{Two-sample t-test results and effect sizes for comparisons between TPE optimization, random search, and the best hyperparameter configuration found by TPE for the MNIST classification task.}
    \label{tab:mnist_tpe_random_two_sample}
\end{table}

\begin{figure}[htbp] 
  \centering 
  \includegraphics[width=\columnwidth]{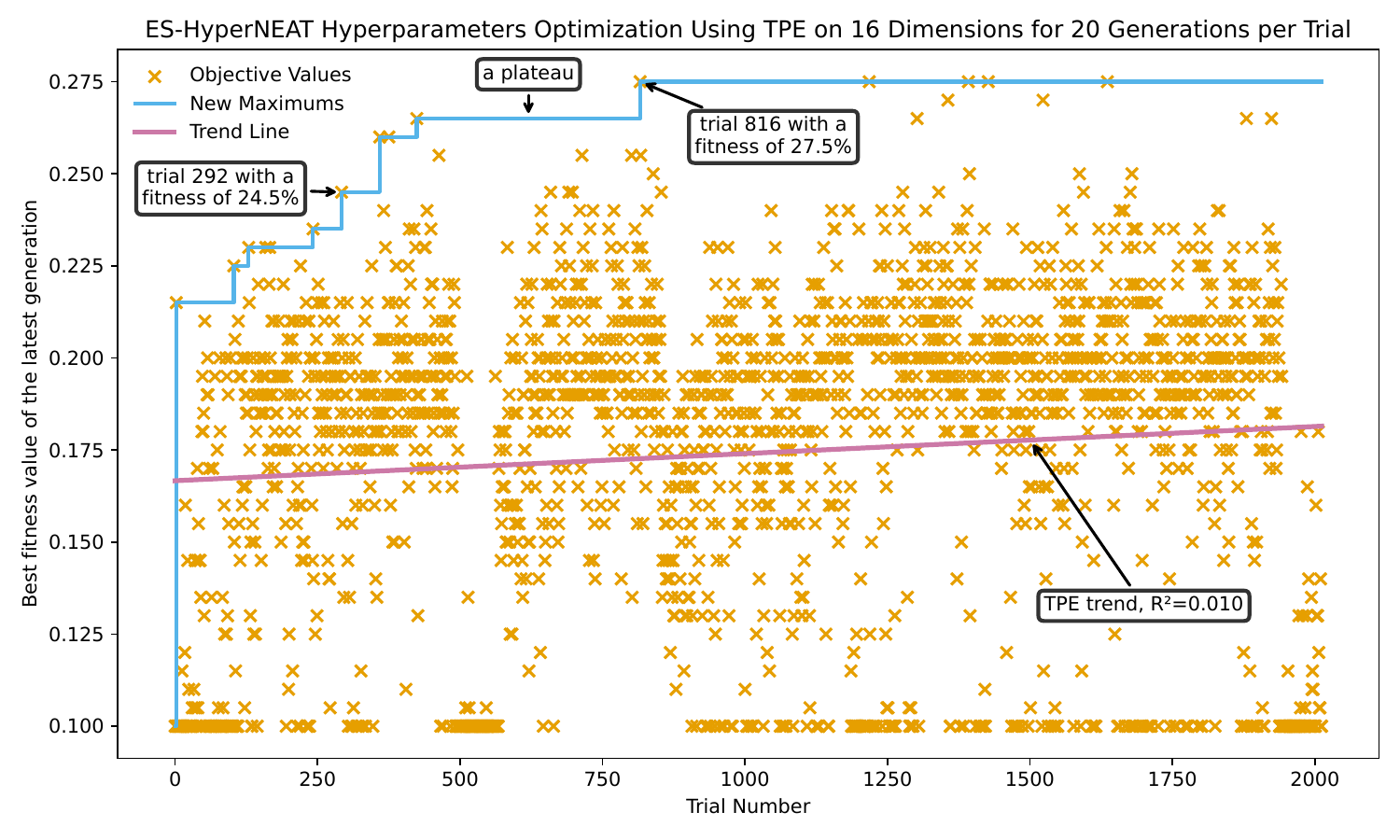} 
  \Description{A scatter plot titled 'ES-HyperNEAT Hyperparameters Optimization Using TPE on 16 Dimensions for 20 Generations per Trial'. The x-axis, labeled 'Trial Number', spans from 0 to 2000 in increments of 250. The y-axis, labeled 'Best fitness value of the latest generation', ranges from 0.10 to 0.275 in increments of 0.025. The plot is composed of numerous orange cross marks indicating the objective values of each trial. Notable features include blue step lines marking new maxima in fitness reached after certain trials, a pink trend line suggesting an overall weak correlation between the trial number and fitness value (indicated by a black arrow pointing to the trend line with the annotation 'TPE trend, R²=0.010'). The trend line shows minimal positive slope. Annotations highlight specific trials, such as 'trial 292 with a fitness of 24.5\%' around the lower center-left of the plot and 'trial 816 with a fitness of 27.5\%' towards the upper center-right. The notation 'a plateau' points to a horizontal collection of data points following a step where the fitness value did not improve over multiple trials. The visualization captures the stochastic nature of the TPE optimization process and iterations where performance improvements were observed.}
  \caption{Scatter plot of the TPE optimization process for the MNIST classification task, showcasing each trial's objective values (best fitness of the latest generation).}
  \label{fig:mnist_tpe_optimization} 
\end{figure}

Figure~\ref{fig:mnist_tpe_optimization} showcases the TPE optimization process for the MNIST classification task, highlighting key points of interest and the stochastic nature of the optimization process. Table~\ref{tab:best_hyperparameters_compact} showcases the hyperparameters that varied across the optimal configurations discovered by TPE, which achieved a 27.5\% accuracy on the MNIST task. The table is divided into NEAT and substrate parameters, with the following hyperparameters remaining constant across all optimal configurations: num\_hidden = 0, pop\_size = 100, initial\_depth = 2, conn\_delete\_prob = 0.5, node\_add\_prob = 0.2, division\_threshold = 0.5, and iteration\_level = 0. The varying hyperparameters, such as connection addition probability (CAP), node deletion probability (NDP), initial connection type (IC), connection fraction (CF), default activation function (AD), maximum depth (MD), variance threshold (VT), maximum weight (MW), and activation function (Act), play a crucial role in determining the network's effectiveness.

\begin{table}[htbp]
    \centering
    \scriptsize
    \begin{tabularx}{\columnwidth}{XXXXX|XXXX}
        \toprule
        \multicolumn{5}{c|}{\textbf{NEAT Parameters}} & \multicolumn{4}{c}{\textbf{Substrate Parameters}} \\
        \textbf{CAP} & \textbf{NDP} & \textbf{IC} & \textbf{CF} & \textbf{AD} & \textbf{MD} & \textbf{VT} & \textbf{MW} & \textbf{Act} \\
        \midrule
        0.5 & 0.8 & FN & 0.8 & softplus & 5 & 0.01 & 3 & clamped \\
        0.5 & 0.2 & FD & 0.4 & clamped  & 4 & 0.03 & 7 & softplus \\
        0.5 & 0.2 & FD & 0.4 & tanh     & 3 & 0.03 & 7 & clamped \\
        0.7 & 0.2 & FD & 0.8 & softplus & 5 & 0.03 & 3 & softplus \\
        \bottomrule
    \end{tabularx}
    \caption{Optimal ES-HyperNEAT hyperparameter configurations discovered by TPE on the MNIST classification task, achieving the best accuracy of 27.5\%.}
    \label{tab:best_hyperparameters_compact}
\end{table}

In summary, our evaluation demonstrates the significant superiority of TPE optimization over random search for hyperparameter tuning in the ES-HyperNEAT algorithm on the MNIST classification task. The substantial improvements in accuracy, the highly significant statistical differences, and the large effect sizes underscore the effectiveness of TPE in finding performant hyperparameter configurations. These findings contribute to the growing body of evidence supporting the use of hyperparameter optimization techniques in neuroevolutionary algorithms, preparing the way for more effective and efficient neural network design and application.

\subsection{Transferability Results}
This subsection presents the results of investigating the generalizability of the best hyperparameter configuration found using TPE search on the MNIST task to logic operations and Fashion-MNIST classification tasks.

\subsubsection{Logic Operations Tasks}

We performed a random search (RandomSearch) over 30 trials and compared the results with the performance of using the logic operation on the best hyperparameter configuration from the MNIST TPE search (MNIST-Config) over 30 trials. We did not extend to 292 trials as we did for the MNIST and Fashion-MNIST experiments because we reached 100\% accuracy within the 30-trial scope. The fitness for each genome is evaluated by calculating the accuracy, over all possible input combinations, of the network’s output compared to the expected output~\cite{stanley2002evolving}.

Table~\ref{tab:logic_ops_accuracy} provides an overview of the performance metrics, comparing the random search (RandomSearch) and the transferred hyperparameter configuration from MNIST TPE Search (MNIST-Config). It presents each logic operation's mean, median, standard deviation, and best and worst accuracy values.

\begin{table}[htbp]
    \centering
    \footnotesize
    \begin{tabularx}{\columnwidth}{lSSSSS}
    \toprule
        Method & {Mean (\%)} & {Median (\%)} & {SD (\%)} & {Best (\%)} & {Worst (\%)} \\
    \midrule
        \multicolumn{6}{l}{\textbf{XOR}} \\
        RandomSearch & 75.19 & 75.27 & 19.28 & 100.00 & 50.00 \\
        MNIST-Config & 83.33 & 83.25 & 1.38 & 87.47 & 79.13 \\
    \midrule
        \multicolumn{6}{l}{\textbf{OR}} \\
        RandomSearch & 76.80 & 98.73 & 30.70 & 100.00 & 25.00 \\
        MNIST-Config & 100.00 & 100.00 & 0.00 & 100.00 & 100.00 \\
    \midrule
        \multicolumn{6}{l}{\textbf{AND}} \\
        RandomSearch & 88.77 & 87.50 & 10.71 & 100.00 & 75.00 \\
        MNIST-Config & 87.91 & 87.50 & 1.11 & 91.15 & 87.41 \\
    \midrule
        \multicolumn{6}{l}{\textbf{NOR}} \\
        RandomSearch & 85.91 & 77.22 & 12.03 & 100.00 & 75.00 \\
        MNIST-Config & 86.65 & 81.17 & 8.56 & 100.00 & 76.01 \\
    \midrule
        \multicolumn{6}{l}{\textbf{XNOR}} \\
        RandomSearch & 73.78 & 73.98 & 20.35 & 100.00 & 50.00 \\
        MNIST-Config & 76.43 & 75.00 & 3.07 & 87.53 & 73.96 \\
    \midrule
        \multicolumn{6}{l}{\textbf{NAND}} \\
        RandomSearch & 66.00 & 80.18 & 31.77 & 100.00 & 25.00 \\
        MNIST-Config & 87.68 & 86.49 & 6.78 & 100.00 & 81.14 \\
    \bottomrule
    \end{tabularx}
    \caption{Accuracy metrics for logic operations tasks, comparing random search (RandomSearch) and the outcome of running the logic operation on the best hyperparameter configuration from the MNIST TPE search (MNIST-Config).}
    \label{tab:logic_ops_accuracy}
\end{table}

The results show that the transferred hyperparameters from MNIST TPE Search (MNIST-Config) outperform RandomSearch for all logic operations, with the most substantial improvements observed for OR, NAND, and XOR. MNIST-Config achieves perfect accuracy (100\%) for the OR operation, while RandomSearch has a mean accuracy of 76.80\%. For NAND, MNIST-Config attains a mean accuracy of 87.68\%, compared to 66\% for RandomSearch. In the case of XOR, MNIST-Config reaches a mean accuracy of 83.33\%, surpassing RandomSearch's 75.19\%.

To quantify the significance of the differences between the methods, we conducted two-sample t-tests assuming unequal variances. Table~\ref{tab:logic_ops_two_sample} presents the t-statistics, p-values, and effect sizes (Cohen's d) for comparing MNIST-Config and RandomSearch for each logic operation.

\begin{table}[htbp]
    \centering
    \footnotesize
    \begin{tabularx}{\columnwidth}{XSSS}
    \toprule
        {MNIST-Config vs. RandomSearch} & {t-statistic} & {p-value} & {Cohen's d} \\
    \midrule
        XOR & 2.31 & 0.025 & 0.60 \\
        OR & 4.14 & {$<$0.001} & 1.07 \\
        AND & -0.43 & 0.663 & -0.11 \\
        NOR & 0.27 & 0.785 & 0.07 \\
        XNOR & 0.71 & 0.483 & 0.18 \\
        NAND & 3.66 & 0.001 & 0.94 \\
    \bottomrule
    \end{tabularx}
    \caption{Two-sample t-test results and effect sizes for comparisons between the outcome of using the best hyperparameter configuration from the MNIST TPE search (MNIST-Config) and random search (RandomSearch) for each logic operation.}
    \label{tab:logic_ops_two_sample}
\end{table}

The results show that the transferred hyperparameters from MNIST TPE Search (MNIST-Config) significantly outperform RandomSearch for the XOR (t = 2.31, p = 0.025, d = 0.60), OR (t = 4.14, p < 0.001, d = 1.07), and NAND (t = 3.66, p = 0.001, d = 0.94) operations, with moderate to large effect sizes. For the AND, NOR, and XNOR operations, the differences between MNIST-Config and RandomSearch are not statistically significant (p > 0.05), with small effect sizes (d < 0.2).

These findings suggest that the best hyperparameter configuration from the MNIST task can be effectively transferred to specific logic operations tasks, particularly XOR, OR, and NAND, leading to significant performance improvements over random search. However, the transferability may be limited for other logic operations, such as AND, NOR, and XNOR, where the performance differences are not statistically significant.

\subsubsection{Fashion-MNIST Classification Task}

We conducted a random search over 30 trials (RandomSearch-30) and compared the results with the performance of using the best hyperparameter configuration from the MNIST TPE search (MNIST-Config) over 30 trials. Table~\ref{tab:fashion_mnist_accuracy} presents the accuracy metrics for the Fashion-MNIST task using RandomSearch-30, RandomSearch-292, and MNIST-Config.

\begin{table}[htbp]
    \centering
    \footnotesize
    \begin{tabularx}{\columnwidth}{lSSSSS}
    \toprule
        {Method} & {Mean (\%)} & {Median (\%)} & {SD (\%)} & {Best (\%)} & {Worst (\%)} \\
    \midrule
        RandomSearch-30 & 11.93 & 10.00 & 3.74 & 23.00 & 10.00 \\
        RandomSearch-292 & 11.60 & 10.00 & 3.08 & 23.50 & 10.00 \\
        MNIST-Config & 20.00 & 20.00 & 1.00 & 22.50 & 18.50 \\
    \bottomrule
    \end{tabularx}
    \caption{Accuracy metrics for the Fashion-MNIST classification task, comparing random searches (RandomSearch-30 and RandomSearch-292) and the use of the best hyperparameter configuration from the MNIST TPE search (MNIST-Config).}
    \label{tab:fashion_mnist_accuracy}
\end{table}

MNIST-Config achieves a mean accuracy of 20.00\%, outperforming RandomSearch-30 (11.93\%) and RandomSearch-292 (11.60\%). However, MNIST-Config attains the lowest best accuracy (22.50\%) compared to RandomSearch-30 (23.00\%) and RandomSearch-292 (23.50\%), suggesting that while the transferred hyperparameters improve overall performance, they may not be optimal for achieving the highest accuracy on the Fashion-MNIST task due to the complexity and differences between the datasets.

Two-sample t-tests assuming unequal variances (tab~\ref{tab:fashion_mnist_two_sample}) reveal no significant difference between RandomSearch-30 and Random\-Search-292 (t = 0.47, p = 0.641, d = 0.10), indicating that increasing the number of trials from 30 to 292 does not substantially improve random search performance for the Fashion-MNIST task. However, MNIST-Config significantly outperforms both RandomSearch-30 (t = 11.42, p < 0.001, d = 2.95) and RandomSearch-292 (t = 32.75, p < 0.001, d = 2.85) with large effect sizes, demonstrating that the best MNIST hyperparameter configuration can be effectively transferred to the Fashion-MNIST classification task, leading to substantial performance improvements over random search.

\begin{table}[htbp]
    \centering
    \footnotesize
    \begin{tabularx}{\columnwidth}{XSSS}
    \toprule
        {Comparison} & {t-statistic} & {p-value} & {Cohen's d} \\
    \midrule
        {RandomSearch-292 vs. RandomSearch-30} & 0.47 & 0.641 & 0.10 \\
        {MNIST-Config vs. RandomSearch-30} & 11.42 & {$<$0.001} & 2.95 \\
        {MNIST-Config vs. RandomSearch-292} & 32.75 & {$<$0.001} & 2.85 \\
    \bottomrule
    \end{tabularx}
    \caption{Two-sample t-test results and effect sizes for the comparisons between applying the best hyperparameter configuration from the MNIST TPE search (MNIST-Config) and random searches (RandomSearch-30 and RandomSearch-292) for the Fashion-MNIST classification task.}
    \label{tab:fashion_mnist_two_sample}
\end{table}
		
In summary, our evaluation of the cross-task transferability of optimized hyperparameters demonstrates that the best MNIST hyperparameter configuration found using TPE search can be effectively transferred to the Fashion-MNIST classification task, leading to significant performance improvements over random search. These findings underscore the potential benefits of leveraging knowledge from one task to improve performance on related tasks, even when the source task (MNIST) is less complex than the target task (Fashion-MNIST). However, further investigation is needed to leverage knowledge from the source task (MNIST), which is more complex than the target tasks (logic operations). The transferability of optimized hyperparameters can save computational resources and time by reducing the need for extensive hyperparameter tuning on each new task. However, the effectiveness of transfer learning may depend on the specific characteristics of the target task and the similarity between the source and target domains.

\subsection{Validation Experiments and Bug Mitigation}

During our experiments, we discovered a bug in our parallel execution code that occasionally caused hyperparameter configurations to be swapped across generations between concurrently running trials. Upon identifying this issue, we promptly halted the experiments, rectified the code, and conducted validation experiments to assess the bug's impact on our findings.

We first examined the validation results for the logic operations and MNIST classification experiments using random search, confirming that the bug did not significantly affect these results. However, due to the substantial computational resources required for the initial TPE experiment, we opted to perform a validation analysis of the TPE results instead of rerunning the entire experiment. This 15-day validation process involved evaluating 885 hyperparameter configurations obtained from the TPE search experiment, with each configuration being run for at least three trials to ensure robustness.

The sequential list of hyperparameter configurations for validation was constructed using a rigorous methodology. We ranked the 2013 configurations from the TPE search based on their objective values, selecting the top 50 configurations and their neighboring configurations within a window of 2. We then randomly sampled 50 additional configurations from the remaining pool, including their neighboring configurations and excluding any preselected configurations. Finally, we incorporated the configurations of 200 randomly selected trials with the lowest objective values, applying the same window and exclusion criteria. This process yielded 271 unique configurations, of which 199 were successfully validated within the allocated timeframe. Concurrently, we allocated additional computational resources to assess the top-performing configurations further, evaluating 8 of the best hyperparameter configurations for 30 trials each, with some experiments prematurely terminated due to resource constraints.

The validation process revealed that 72 out of the 199 assessed configurations surpassed the best result obtained by the corrected random search (22.00\%), providing persuasive evidence for TPE optimization's superiority over random search in the MNIST classification experiment using ES-HyperNEAT, even with the hyperparameter swapping bug present.

Table~\ref{tab:corrected_accuracies} presents the mean, median, standard deviation, best, and worst accuracy values for the MNIST, Logic Operations, Fashion-MNIST classification tasks for the corrected random search over 30 trials (RS-Corr-30), the corrected random search of 292 trials (RS-Corr-292), the validation of 199 hyperparameter configurations from the MNIST TPE search (TPE-Validation) and the best hyperparameter configurations found during the validation of the MNIST TPE search (TPE-Val-Best) applied to the different tasks.

\begin{table}[htbp]
  \centering
  \footnotesize
  \begin{tabularx}{\columnwidth}{lSSSSS}
  \toprule
    {Method} & {Mean (\%)} & {Median (\%)} & {SD (\%)} & {Best (\%)} & {Worst (\%)} \\
  \midrule
    \multicolumn{6}{l}{\textbf{MNIST}} \\
    RS-Corr-30 & 11.13 & 10.00 & 2.62 & 19.50 & 10.00 \\
    RS-Corr-292 & 11.18 & 10.00 & 2.62 & 22.00 & 10.00 \\
    TPE-Validation & 16.68 & 18.50 & 5.01 & 29.00 & 10.00 \\
    TPE-Val-Best & 20.95 & 20.75 & 2.41 & 29.00 & 17.00 \\
  \midrule
    \multicolumn{6}{l}{\textbf{OR}} \\
    RS-Corr-30 & 60.11 & 55.21 & 35.73 & 100.00 & 25.00 \\
    TPE-Val-Best & 99.92 & 100.00 & 0.33 & 100.00 & 98.26 \\
  \midrule
    \multicolumn{6}{l}{\textbf{AND}} \\
    RS-Corr-30 & 79.21 & 75.00 & 8.11 & 100.00 & 75.00 \\
    TPE-Val-Best & 86.54 & 87.05 & 1.12 & 87.50 & 83.87 \\
  \midrule
    \multicolumn{6}{l}{\textbf{XOR}} \\
    RS-Corr-30 & 58.43 & 50.00 & 14.18 & 100.00 & 50.00 \\
    TPE-Val-Best & 78.34 & 77.61 & 3.33 & 83.29 & 75.00 \\
  \midrule
    \multicolumn{6}{l}{\textbf{NAND}} \\
    RS-Corr-30 & 56.35 & 75.00 & 30.40 & 100.00 & 20.00 \\
    TPE-Val-Best & 86.99 & 85.20 & 6.08 & 99.47 & 80.59 \\
  \midrule
    \multicolumn{6}{l}{\textbf{NOR}} \\
    RS-Corr-30 & 77.13 & 75.00 & 6.50 & 100.00 & 75.00 \\
    TPE-Val-Best & 84.82 & 81.25 & 8.62 & 100.00 & 75.00 \\
  \midrule
    \multicolumn{6}{l}{\textbf{XNOR}} \\
    RS-Corr-30 & 57.67 & 50.00 & 13.19 & 100.00 & 50.00 \\
    TPE-Val-Best & 74.32 & 74.75 & 1.97 & 80.33 & 69.35 \\
  \midrule
    \multicolumn{6}{l}{\textbf{Fashion-MNIST}} \\
    TPE-Val-Best & 20.77 & 20.25 & 1.34 & 24.00 & 19.00 \\
  \bottomrule
  \end{tabularx}
  \caption{Accuracy metrics for MNIST, logic operations tasks (OR, AND, XOR, NAND, NOR, XNOR), and Fashion-MNIST, using correct random search (RS-Corr-30 and RS-Corr-292), Validation of TPE Search on MNIST (TPE-Validation), and the best configurations found during the validation of the MNIST TPE search (TPE-Val-Best) applied to the different tasks.}
  \label{tab:corrected_accuracies}
\end{table}

Statistical analysis using two-sample t-tests (tab~\ref{tab:validation_two_sample}) showed no significant differences between the original and corrected random searches for the MNIST task, suggesting the bug's minimal impact on the random search results and providing a reliable baseline for comparison with TPE optimization. In contrast, TPE-Val-Best significantly outperformed RS-Corr-292 for the MNIST task (t = 20.93, p < 0.001, d = 3.72), providing strong evidence for TPE optimization's superiority over random search.

\begin{table}[htbp]
    \centering
    \footnotesize
    \begin{tabularx}{\columnwidth}{lSSS}
        \toprule
        {Comparison} & {t-statistic} & {p-value} & {Cohen's d} \\
        \midrule
        \multicolumn{4}{l}{\textbf{MNIST}} \\
        RandomSearch-30 vs. RS-Corr-30 & 0.48 & 0.632 & 0.12 \\
        RandomSearch-292 vs. RS-Corr-292 & 1.01 & 0.311 & 0.08 \\
        TPE-Search vs. RS-Corr-30 & 12.76 & {$<$0.001} & 1.49 \\
        TPE-Search vs. RS-Corr-292 & 34.34 & {$<$0.001} & 1.53 \\
        TPE-Val-Best vs. RS-Corr-30 & 15.77 & {$<$0.001} & 4.07 \\
        TPE-Val-Best vs. RS-Corr-292 & 20.93 & {$<$0.001} & 3.72 \\
        TPE-Val-Best vs. TPE-Search & 7.86 & {$<$0.001} & 0.84 \\
        \midrule
        \multicolumn{4}{l}{\textbf{OR}} \\
        RandomSearch-30 vs. RS-Corr-30 & 1.94 & 0.057 & 0.50 \\
        TPE-Val-Best vs. RS-Corr-30 & 6.10 & {$<$0.001} & 1.58 \\
        \midrule
        \multicolumn{4}{l}{\textbf{AND}} \\
        RandomSearch-30 vs. RS-Corr-30 & 3.89 & {$<$0.001} & 1.00 \\
        TPE-Val-Best vs. RS-Corr-30 & 4.90 & {$<$0.001} & 1.27 \\
        \midrule
        \multicolumn{4}{l}{\textbf{XOR}} \\
        RandomSearch-30 vs. RS-Corr-30 & 3.83 & {$<$0.001} & 0.99 \\
        TPE-Val-Best vs. RS-Corr-30 & 7.49 & {$<$0.001} & 1.93 \\
        \midrule
        \multicolumn{4}{l}{\textbf{NOR}} \\
        RandomSearch-30 vs. RS-Corr-30 & 3.52 & 0.001 & 0.91 \\
        TPE-Val-Best vs. RS-Corr-30 & 3.91 & {$<$0.001} & 1.01 \\
        \midrule
        \multicolumn{4}{l}{\textbf{NAND}} \\
        RandomSearch-30 vs. RS-Corr-30 & 1.20 & 0.234 & 0.31 \\
        TPE-Val-Best vs. RS-Corr-30 & 5.41 & {$<$0.001} & 1.40 \\
        \midrule
        \multicolumn{4}{l}{\textbf{XNOR}} \\
        RandomSearch-30 vs. RS-Corr-30 & 3.64 & 0.001 & 0.94 \\
        TPE-Val-Best vs. RS-Corr-30 & 6.84 & {$<$0.001} & 1.76 \\
        \midrule
        \multicolumn{4}{l}{\textbf{Fashion-MNIST}} \\
        TPE-Val-Best vs. RandomSearch-30 & 12.19 & {$<$0.001} & 3.15 \\
        TPE-Val-Best vs. RandomSearch-292 & 30.13 & {$<$0.001} & 3.09 \\
        \bottomrule
    \end{tabularx}
    \caption{Two-sample t-test results and effect sizes for key comparisons in the validation experiments and bug mitigation process.}
    \label{tab:validation_two_sample}
\end{table}

The consistency of the results across multiple comparisons and the substantial effect sizes demonstrate the robustness of TPE optimization's superiority and the effectiveness of the best hyperparameter configurations, emphasizing the importance of thorough experimental validation and careful design, execution, and monitoring of experiments in research.

\section{Discussion}

\subsection{Hyperparameter Optimization}
This study demonstrates the effectiveness of TPE for optimizing ES-HyperNEAT hyperparameters on the MNIST classification task, consistently outperforming random search. The logic operations experiments suggest that these tasks are relatively easy for ES-HyperNEAT, and extensive hyperparameter optimization may not be necessary. The transferability of hyperparameters from MNIST to logic operations tasks shows promise, with the best MNIST configurations consistently outperforming corrected random searches.

The Fashion-MNIST classification task provides preliminary evidence of the best MNIST hyperparameter configuration's transferability to a similar task. However, the extent of these improvements may be limited due to inherent differences between the datasets, such as Fashion-MNIST's increased complexity and diversity, potential overspecialization of the MNIST configuration, and the need for task-specific input encoding and output decoding strategies.

\subsection{Implications of Transferability Results}
The transferability experiments provide valuable insights into the generalizability of optimized hyperparameters across different tasks. The success of transferring the best MNIST configuration to certain logic operations tasks (XOR, OR, and NAND) may be due to the relative simplicity of these problems, as they can be solved using random search with just 30 trials. The limited transferability observed to other logic operations (AND, NOR, and XNOR) further supports this interpretation, suggesting that the success of transfer learning in these cases may depend on the specific characteristics of the target task and the luck of the draw in terms of the hyperparameter configuration.

The transferability results for the Fashion-MNIST classification task offer more substantial evidence supporting the potential advantages of transfer learning in neuroevolution. During the TPE validation process, the best hyperparameter configurations discovered for the MNIST task, referred to as TPE-Val-Best in Table~\ref{tab:corrected_accuracies}, achieved a mean accuracy of 20.77\% when applied to the Fashion-MNIST dataset. This performance surpasses that of RandomSearch-30 (11.93\%), as shown in Table~\ref{tab:fashion_mnist_accuracy}. These findings suggest that transferring optimized hyperparameters from a source task to a related target task sharing similar characteristics could enhance performance compared to conducting a random search, potentially circumventing the need for extensive computational resources.

These findings highlight the importance of considering the relationship between the source and target tasks when applying transfer learning techniques in neuroevolution. While transferring optimized hyperparameters may provide performance benefits when the tasks are closely related, the effectiveness of this approach may be limited when the problem domains differ substantially or when the target task is relatively simple.

\subsection{Robustness of Results}

The bug's discovery in our parallel execution code underscores the importance of diligent validation and bug resolution in experimental research. The validation results revealed that the bug significantly affected the random search performance for the AND, XOR, NOR, and XNOR tasks, leading to an overestimation of the baseline performance with p $\le$ 0.001 (tab~\ref{tab:validation_two_sample}). However, the bug did not significantly impact the random search results for the MNIST, OR, NAND, and Fashion-MNIST tasks, as the Fashion-MNIST random search was performed after the bug was fixed, providing reliable baselines for comparison with TPE optimization.

The rigorous validation process, which evaluated 885 hyperparameter configurations from the TPE search experiment, ensured the robustness of the results. The finding that 72 assessed configurations surpassed the best result obtained by the corrected random search (22.00\%) demonstrates the superiority of TPE optimization over random search, even in the bug's presence.

The statistical analysis confirmed the significance of the observed differences, with highly significant results and large effect sizes between the best validated MNIST configuration (TPE-Val-Best) and the corrected random search (RS-Corr-30 and RS-Corr-292). These findings highlight the importance of thorough experimental validation and careful design, execution, and monitoring of experiments.

The bug mitigation process and subsequent validation experiments demonstrate the resilience of our findings and the robustness of the TPE optimization approach for the MNIST, OR, NAND, and Fashion-MNIST tasks. However, they also emphasize the need for caution when interpreting results for the AND, XOR, NOR, and XNOR tasks, as the bug significantly affected the random search performance, leading to an overestimation of the baseline performance. Despite this, the ability of random search to find optimal solutions within 30 trials suggests that these tasks may be relatively simple for ES-HyperNEAT, and the consistent outperformance of the transferred best MNIST configuration (TPE-Val-Best) highlights the potential benefits of using optimized hyperparameters from a more complex task, even for simpler problems.

\subsection{Limitations and Future Work}
Our study provides valuable insights into hyperparameter optimization for ES-HyperNEAT and the transferability of optimized hyperparameters. However, it also reveals several limitations that present opportunities for future research, particularly in exploring alternative optimization algorithms and expanding the search space to enhance the efficiency and effectiveness of neuroevolutionary algorithms.

One of our primary objectives for future work is to investigate optimization algorithms, mainly based on evolutionary computation, such as CMA-ES, which could potentially improve the hyperparameter optimization process for ES-HyperNEAT. By comparing the performance of these algorithms with TPE, we aim to identify the most suitable approach for efficiently optimizing neuroevolutionary algorithms.

Furthermore, we plan to expand the hyperparameter search space by increasing the granularity of the hyperparameters, modifying the substrate geometric initialization, and exploring additional hidden layers and higher population sizes. This expansion will enable us to identify the most influential hyperparameters through sensitivity analysis and potentially overcome limitations in achievable fitness on complex tasks like MNIST classification.

Another crucial aspect of our future research is to delve deeper into the transferability of optimized hyperparameters across tasks with varying complexity. By conducting extensive experiments on a broader range of tasks from domains such as games and the discovery of statistical models, we aim to develop a strong understanding of how transferability works in practice. This knowledge will serve as a foundation for bootstrapping new tasks efficiently and effectively, ultimately contributing to developing more adaptable and high-performing neural networks. To achieve these goals, we will also focus on developing methods for identifying the most relevant source tasks for a given target domain and adapting transferred hyperparameters to the specific requirements of the target task. Exploring advanced transfer learning approaches, such as meta-learning or multi-task learning, may further improve the generalizability of evolved networks across a wide range of tasks and domains.

\section{Conclusion}

This study represents a significant step forward in the field of neuroevolution, demonstrating the effectiveness of the Tree-structured Parzen Estimator (TPE) for optimizing the hyperparameters of the Evolvable-Substrate HyperNEAT (ES-HyperNEAT) algorithm on the MNIST classification task. The findings highlight the superiority of TPE over random search in identifying performant hyperparameter configurations, with the best configuration discovered by TPE achieving a noteworthy accuracy of 29.00\% on MNIST (tab~\ref{tab:corrected_accuracies}) using a significantly smaller population size and fewer generations compared to previous studies.

The cross-task transferability experiments provide valuable insights into the generalizability of optimized hyperparameters. The results suggest that the best MNIST configuration can be effectively transferred to the Fashion-MNIST classification task, a more complex problem sharing similar characteristics with MNIST, leading to significant improvements over random search. However, the transferability to simpler tasks, such as certain logic operations, is less conclusive. These findings highlight that the effectiveness of transferring optimized hyperparameters may depend on the complexity and similarity of the problem domains.

This research introduces novel perspectives on hyperparameter optimization and transferability in neuroevolution, underscoring the importance of rigorous experimental validation. The extensive validation process, which evaluated a wide range of hyperparameter configurations obtained from the TPE search experiment, demonstrates the resilience of the findings and the robustness of the TPE optimization approach. The consistency of the results across multiple comparisons and the substantial effect sizes emphasize the reliability and reproducibility of the research outcomes, highlighting the significance of meticulous experimental practices in ensuring the integrity of scientific findings.

The findings have meaningful implications for the practical application of neuroevolutionary algorithms, laying the groundwork for developing more efficient and adaptable neural networks. By addressing the limitations and opportunities identified in this study, future research can focus on exploring alternative optimization algorithms, expanding the search space, and delving deeper into the transferability of optimized hyperparameters across tasks with varying complexity.

In conclusion, this study contributes significantly to neuroevolution, demonstrating the potential of hyperparameter optimization techniques like TPE to enhance the performance of algorithms such as ES-HyperNEAT and highlighting the importance of rigorous experimental practices and validation. The insights gained from this research lead the way for unlocking the full potential of neuroevolutionary algorithms, ultimately contributing to advancing the field and realizing more efficient, adaptable, and high-performing neural networks.

\begin{acks}
This research was supported by the Doc.Mobility scholarship awar\-ded by the University of Neuchâtel, Switzerland, and swissuniversities.
\end{acks}

\bibliographystyle{ACM-Reference-Format}
\bibliography{references}

@String{Computing = "Computing" }

@String{Computer = "{IEEE} Computer" }

@String{Springer = "Springer-Verlag" }

@article{stanley2002evolving,
  title={Evolving neural networks through augmenting topologies},
  author={Stanley, Kenneth O and Miikkulainen, Risto},
  journal={Evolutionary computation},
  volume={10},
  number={2},
  pages={99--127},
  year={2002},
  publisher={MIT Press}
}

@article{stanley2009hypercube,
  title={A hypercube-based encoding for evolving large-scale neural networks},
  author={Stanley, Kenneth O and D'Ambrosio, David B and Gauci, Jason},
  journal={Artificial life},
  volume={15},
  number={2},
  pages={185--212},
  year={2009},
  publisher={MIT Press One Rogers Street, Cambridge, MA 02142-1209, USA journals-info~…}
}

@inproceedings{bergstra2011algorithms,
 author = {Bergstra, James and Bardenet, R\'{e}mi and Bengio, Yoshua and K\'{e}gl, Bal\'{a}zs},
 booktitle = {Advances in Neural Information Processing Systems},
 editor = {J. Shawe-Taylor and R. Zemel and P. Bartlett and F. Pereira and K.Q. Weinberger},
 pages = {},
 publisher = {Curran Associates, Inc.},
 title = {Algorithms for Hyper-Parameter Optimization},
 url = {https://proceedings.neurips.cc/paper_files/paper/2011/file/86e8f7ab32cfd12577bc2619bc635690-Paper.pdf},
 volume = {24},
 year = {2011},
 email = {}
}

@article{stanley2007compositional,
  title={Compositional pattern producing networks: A novel abstraction of development},
  author={Stanley, Kenneth O},
  journal={Genetic programming and evolvable machines},
  volume={8},
  pages={131--162},
  year={2007},
  publisher={Springer}
}

@article{snoek2012practical,
  title={Practical bayesian optimization of machine learning algorithms},
  author={Snoek, Jasper and Larochelle, Hugo and Adams, Ryan P},
  journal={Advances in neural information processing systems},
  volume={25},
  year={2012}
}

@article{shahriari2015taking,
  title={Taking the human out of the loop: A review of Bayesian optimization},
  author={Shahriari, Bobak and Swersky, Kevin and Wang, Ziyu and Adams, Ryan P and De Freitas, Nando},
  journal={Proceedings of the IEEE},
  volume={104},
  number={1},
  pages={148--175},
  year={2015},
  publisher={IEEE}
}

@software{westh2017pureples,
  author = {Westh et al.},
  title = {Pureples - pure python library for es-hyperneat},
  year = {2017},
  publisher = {GitHub},
  journal = {GitHub repository},
  howpublished = {\url{https://github.com/ukuleleplayer/pureples}},
  commit = {312f6c7c6f6dc8352365fd6543725cc6b935a1f6}
}

@inproceedings{akiba2019optuna,
author = {Akiba, Takuya and Sano, Shotaro and Yanase, Toshihiko and Ohta, Takeru and Koyama, Masanori},
title = {Optuna: A Next-generation Hyperparameter Optimization Framework},
year = {2019},
isbn = {9781450362016},
publisher = {Association for Computing Machinery},
address = {New York, NY, USA},
url = {https://doi.org/10.1145/3292500.3330701},
doi = {10.1145/3292500.3330701},
booktitle = {Proceedings of the 25th ACM SIGKDD International Conference on Knowledge Discovery \& Data Mining},
pages = {2623–2631},
numpages = {9},
location = {Anchorage, AK, USA},
series = {KDD '19}
}

@article{bergstra2012random,
  title={Random search for hyper-parameter optimization.},
  author={Bergstra, James and Bengio, Yoshua},
  journal={Journal of machine learning research},
  volume={13},
  number={2},
  year={2012}
}

@article{risi2012enhanced,
  title={An enhanced hypercube-based encoding for evolving the placement, density, and connectivity of neurons},
  author={Risi, Sebastian and Stanley, Kenneth O},
  journal={Artificial life},
  volume={18},
  number={4},
  pages={331--363},
  year={2012},
  publisher={MIT Press 55 Hayward Street, Cambridge, MA 02142-1315, USA}
}

@inproceedings{risi2010indirectly,
  title={Indirectly encoding neural plasticity as a pattern of local rules},
  author={Risi, Sebastian and Stanley, Kenneth O},
  booktitle={International conference on simulation of adaptive behavior},
  pages={533--543},
  year={2010},
  organization={Springer}
}

@article{lecun1998mnist,
  title={The MNIST database of handwritten digits},
  author={LeCun, Yann},
  journal={http://yann. lecun. com/exdb/mnist/},
  year={1998}
}

@article{xiao2017fashion,
  title={Fashion-mnist: a novel image dataset for benchmarking machine learning algorithms},
  author={Xiao, Han and Rasul, Kashif and Vollgraf, Roland},
  journal={arXiv preprint arXiv:1708.07747},
  year={2017}
}

@article{welch1947generalization,
  title={The generalization of ‘STUDENT'S’ problem when several different population variances are involved},
  author={Welch, Bernard L},
  journal={Biometrika},
  volume={34},
  number={1-2},
  pages={28--35},
  year={1947},
  publisher={Oxford University Press}
}

@book{cohen2013statistical,
  title={Statistical power analysis for the behavioral sciences},
  author={Cohen, Jacob},
  year={2013},
  publisher={Routledge}
}

@software{McIntyre_neat-python,
author = {McIntyre, Alan and Kallada, Matt and Miguel, Cesar G. and Feher de Silva, Carolina and Netto, Marcio Lobo},
title = {{neat-python}}
}

@article{papavasileiou2021systematic,
  title={A systematic literature review of the successors of “neuroevolution of augmenting topologies”},
  author={Papavasileiou, Evgenia and Cornelis, Jan and Jansen, Bart},
  journal={Evolutionary computation},
  volume={29},
  number={1},
  pages={1--73},
  year={2021},
  publisher={MIT Press}
}

@book{devore2016probability,
  address = {Boston, MA},
  author = {Devore, Jay L.},
  edition = {Ninth},
  isbn = {978-1-305-25180-9},
  publisher = {Cengage Learning},
  title = {Probability and Statistics for Engineering and the Sciences},
  year = 2016
}

@article{frazier2018tutorial,
  title={A tutorial on Bayesian optimization},
  author={Frazier, Peter I},
  journal={arXiv preprint arXiv:1807.02811},
  year={2018}
}

@inproceedings{verbancsics2015image,
  title={Image classification using generative neuro evolution for deep learning},
  author={Verbancsics, Phillip and Harguess, Josh},
  booktitle={2015 IEEE winter conference on applications of computer vision},
  pages={488--493},
  year={2015},
  organization={IEEE}
}

@article{lecun1998gradient,
  title={Gradient-based learning applied to document recognition},
  author={LeCun, Yann and Bottou, L{\'e}on and Bengio, Yoshua and Haffner, Patrick},
  journal={Proceedings of the IEEE},
  volume={86},
  number={11},
  pages={2278--2324},
  year={1998},
  publisher={Ieee}
}
\end{document}